\documentclass[a4paper]{article}

\usepackage{INTERSPEECH2018}
\usepackage{multirow}
\usepackage{comment}

\title{Spoken SQuAD: A Study of Mitigating the Impact of \\ Speech Recognition Errors on Listening Comprehension}
\name{Chia-Hsuan Li,Szu-Lin Wu,Chi-Liang Liu,Hung-yi Lee}
\address{College of Electrical Engineering and Computer Science, National Taiwan University, Taiwan}
\email{\{chiahsuan.li,riviera1020,liangtaiwan1230,tlkagkb93901106\}@gmail.com}

\begin{document}
\maketitle
\begin{abstract}
Reading comprehension has been widely studied. One of the most representative reading comprehension tasks is Stanford Question Answering Dataset (SQuAD), on which machine is already comparable with human. 
On the other hand, accessing large collections of  multimedia or spoken content is much more difficult and time-consuming than  plain text content for humans. It's therefore highly attractive to develop machines which can automatically understand spoken content.
In this paper, we propose a new listening comprehension task -– Spoken SQuAD. On the new task, we found that speech recognition errors have catastrophic impact on machine comprehension, and several approaches are proposed to mitigate the impact.
\end{abstract} 

\section{Introduction}
Machine comprehension (MC) on text has significant progress in the recent years.
On Stanford Question Answering Dataset (SQuAD)~\cite{rajpurkar2016squad}, deep neural network (DNN) based models are  comparable with human. 
The achievements of the state-of-the-art models demonstrate that MC models already contain decent reasoning ability. 
On the other hand, accessing large collections of  multimedia or spoken content is much more difficult and time-consuming than  plain text content for humans. 
It is therefore highly attractive to develop Spoken Question Answering (SQA)~\cite{ispoken,comas2012factoid,turmo2008overview,comas2012sibyl}, which requires machine to find the answer from spoken content given a question in either text or spoken form. 
We focus on text query in the following discussions.

Comparing to the significant progress in MC on text documents, MC on spoken content is a much less investigated field.
Different kinds of DNN systems have been used in slot filling task including Recurrent Neural Network (RNN)~\cite{mesnil2015using,simonnet2015exploring}, bidirectional RNN~\cite{hakkani2016multi} and Convolutional Neural Network(CNN)~\cite{vu2016sequential,deoras2013joint}.
However, all these previous models work on sequence labeling task, while SQA requires machine to perform more sophisticated reasoning. 
There were also works trying to estimate word confidence scores or error probability on ASR transcriptions~\cite{hakkani2006beyond,tur2002improving,yu2016abstractive,shiang2014spoken,ghannay2016acoustic,ghannay2015combining,tam2014asr,chen2013asr,zukerman2017improving,bechet2013asr}.
ASR confidence measures have been introduced as additional features for spoken language understanding (SLU)~\cite{simonnet2017asr}.

SQA usually worked with automatic speech recognition (ASR) transcripts of the spoken content, and typical approaches used information retrieval (IR) techniques ~\cite{shiang2014spoken} or used knowledge bases~\cite{hixon2015learning} to find the proper answer. 
Recently, deep learning is used to answer TOEFL listening comprehension test which is a task highly related to SQA.
TOEFL is an English examination that tests the knowledge and skills of academic English for English learners whose native languages is not English. 
On this task,  attention-based RNN is used to construct sentence representation considering the word order~\cite{tseng2016towards}, and tree-structured RNN is used to construct sentence representation  from their syntactic structure~\cite{fang2016hierarchical}.
However, the scale of TOEFL dataset used in the previous study is too limited to develop powerfully expressive models. 
In addition, the test is multi-select, in which machine does not provide an answer by itself, so it is still one step away from SQA.

To further push the boundary of MC on  spoken content,  we propose a new listening comprehension task -– Spoken SQuAD.
The contributions of our work are three-folds:
\begin{itemize}
\item We propose the Spoken SQuAD task, which is an extraction-based SQA task, together with a new evaluation approach. 
\item We found that ASR errors have catastrophic impact on QA.  We tested numbers of state-of-the-art SQuAD models on Spoken SQuAD dataset and reported their degrading performance on ASR transcriptions. 
\item Last but not least, we propose several approaches to mitigate the impact of ASR errors. 
\end{itemize}


\begin{figure}[t]
  \centering
  \includegraphics[width=\linewidth]{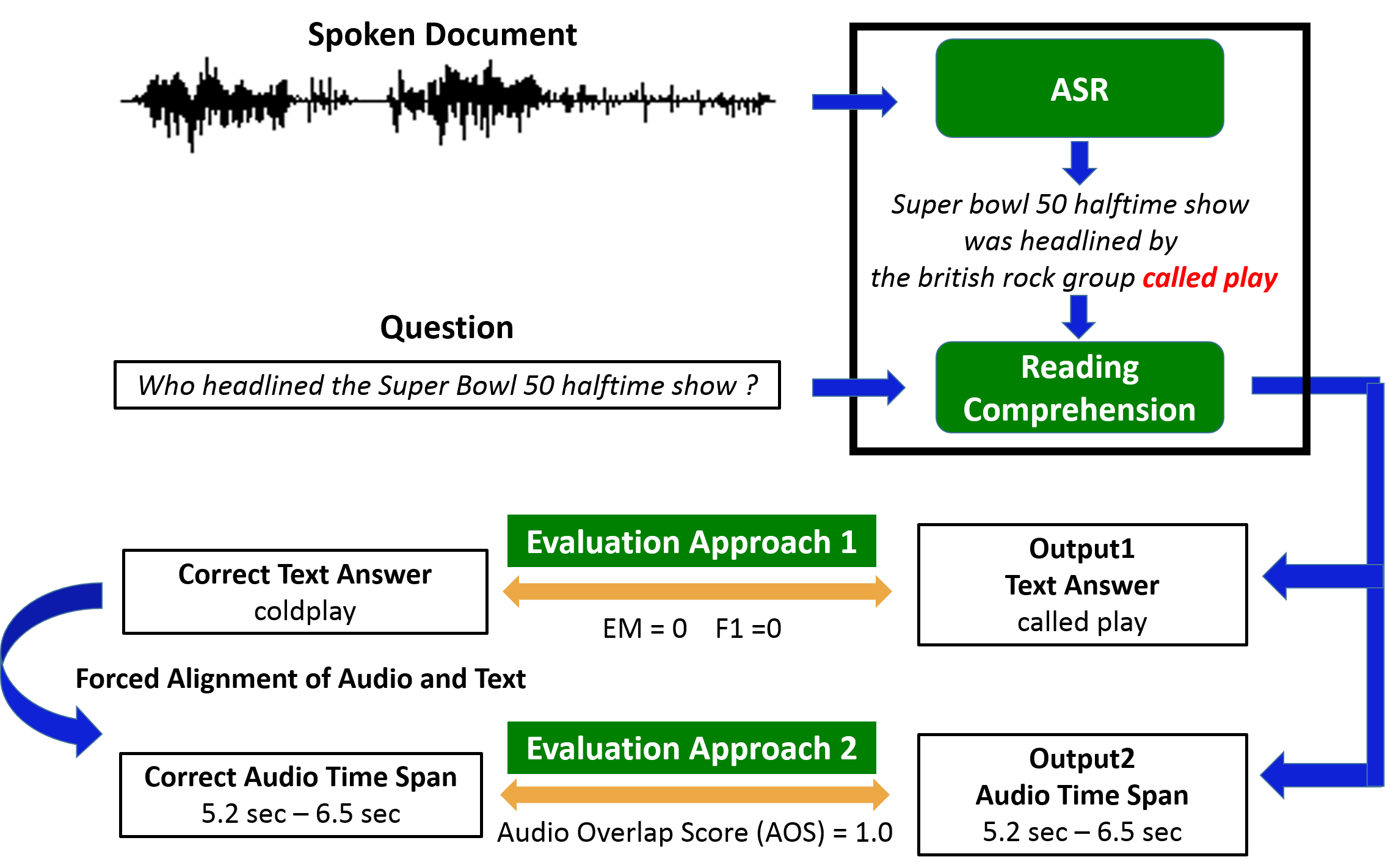}
  \caption{Flow diagram of the SQA and two evaluation methods. 
  Given spoken document and text question, SQA system can have two types  of outputs: a text answer or a time span of audio segments including the answer.
The two types  of outputs will be evaluated by two different approaches.}
  \label{fig:sqa_task}
\end{figure}

\section{Task Definition} 

\subsection{Corpus Description} 
In this paper, we   propose a new listening comprehension task   named Spoken SQuAD, in which document is in spoken form, and the input question is in the form of text. 
SQuAD is one of the largest MC dataset on a large set of Wikipedia articles, with more than 100,000 human-written questions. The answer to each question is always a span in the document. The authors randomly partitioned articles into a training set (80\%), a development set (10\%), and a testing set (10\%). The testing set is not yet publicly available. 
To build a spoken version SQuAD dataset, we conducted the following procedures to generate spoken documents from the original SQuAD dataset. 
First, we used Google text-to-speech system to generate the spoken version of the articles in SQuAD. 
Then we utilized CMU Sphinx~\cite{walker2004sphinx} to generate the corresponding ASR transcriptions. 
In this study, we left the questions in the  text form.
We used SQuAD training set to generate the training set of Spoken SQuAD, and SQuAD development set was used to generate the testing set for Spoken SQuAD. 
If the answer of a question did not exist in the ASR transcriptions of the associated article, we removed  the question-answer pair from the dataset because these examples are too difficult for listening comprehension machine at this stage. 
In this way, we collected 37,111 question answer pairs as the training set and 5,351 as the testing set. 
The WER on the training and testing sets are 22.77\% and 22.73\%, respectively\footnote{Spoken SQuAD dataset : https://github.com/chiahsuan156/Spoken-SQuAD}.

To test the comprehension ability of machine in real life scenario under worse audio quality, we further added two different levels of white noise into the audio files of testing set to obtain different WERs.
The resulting three versions of testing sets and corresponding WERs are listed in Table~\ref{tab:WER}.
The synthesized speech  makes the task easier than its real application, but in the following experiments, we found that the comprehension capability already seriously degraded in the above scenario.
We leave the study on real speech recording as the future work.



\begin{table}[]
\centering
\caption{Three  testing sets with different levels of WER.}
\label{tab:WER}
\begin{tabular}{|c|c|c|c|}
 \hline
 \textbf{Testing Set} &  \textbf{No noise} &\textbf{Noise V1 }&\textbf{Noise V2}\\
 \hline
 \hline
 WER(\%) & 22.73 & 44.22 & 54.82 \\

\hline
\end{tabular}
\end{table}

\subsection{Evaluation  Metrics} 
In this task, when the model is given a spoken article, it needs to find the answer of a text-formed  question. 
SQA can be solved by the concatenation of ASR module and reading comprehension module.
The flow diagram of spoken SQuAD is illustrated in Figure~\ref{fig:sqa_task}.
With the ASR transcriptions of spoken documents, the reading comprehension module can return a text answer based on the questions.
The most intuitive way to evaluate the text answer is to directly compute the Exact Match (EM) and F1 scores between the predicted text answer and the ground-truth text answer.
If the predicted text answer and the ground-truth text answer are exactly the same, then the EM score is 1, otherwise 0. 
The F1 score is based on the precision and recall.
Precision   is the percentage of words in the predicted answer existing in the ground-truth answer, while the recall  is the percentage of words in the ground-truth answer also appearing in the predicted answer. 

The above evaluation on text answer not only considers the performance of the reading comprehension system but also the ASR system. 
If the reading comprehension system correctly identifies the word sequence in the transcription that corresponds to the answer, but the words in the answers are misrecognized, then the answer would be considered as completely incorrect. 
This is very possible because most answers are name entities which are usually Out-of-Vocabulary.
Therefore, we propose a second evaluation approach specifically designed for SQA.
In the second approach, instead of returning a text answer, machine returns the time span corresponding to the answer.
In other words, it outputs an audio segment extracted from spoken document as the answer. 
The time span of the correct answer in the spoken document is  available.
Because in SQuAD a correct answer always exists in the document, we can force-align the spoken document with its reference transcription to obtain the time span of the correct answer. 
We compute Audio Overlapping Score (AOS) between the  time span of system output and the ground-truth time span as in (\ref{eq1}).
\begin{equation}
AOS = \frac{X \cap Y}{X \cup Y},
  \label{eq1}
\end{equation}
where $X$ is the audio time interval of  predicted answer, and $Y$ is the audio time interval of ground-truth answer. 
AOS rewards the high portion of overlapping and punishes the model for predicting too long time spans. 
The second evaluation approach is necessary.
In a listening comprehension test, it is extremely difficult, even for human, to correctly transcribe every single word in the spoken document. 
However, if one comprehends the whole spoken document in the right way, he/she can easily select the right segments corresponding to the answer in the audio file.
Therefore, we believe the second approach better evaluates the reasoning capability of SQA system. 
In the following experiments, because the models always select the answers from the ASR transcriptions, to compute AOS, we simply used time stamp for each word given by ASR module to find the time span of the predicted answers.
We believe in the future an end-to-end SQA system can probably directly output time span without generating a text answer first.


\section{Model}

The SQA system is the cascade of an ASR module and reading comprehension module. 
In this section, we first briefly introduce the reading comprehension models we used. 
Then we introduce how we used subword sequence embedding  to mitigate the impact of ASR errors.

\subsection{Reading Comprehension Models}

We chose several competitive models that have acquired state-of-the-art results on SQuAD.The models  are listed as follow:
\begin{itemize}
\item \textbf{BiDirectional Attention Flow (BiDAF)}~\cite{seo2016bidirectional}

  \item \textbf{R-NET}~\cite{wang2017gated}
  
  \item \textbf{Mnemonic Reader}~\cite{hu2017reinforced}
  \item \textbf{FusionNet}~\cite{huang2017fusionnet}
  \item \textbf{Dr.QA}~\cite{chen2017reading}
  \end{itemize}
In our task, during the testing, those models take a machine-transcribed spoken document and a question as input, and the output is an extracted text span from the ASR transcription. 
We first train these models on SQuAD training set and compare the testing results between SQuAD dev set and Spoken SQuAD testing set.
Furthermore, the models can be trained on text documents or the ASR transcriptions of spoken documents. 
We will also compare the results from the two kinds of training data.

\subsection{Subword Units}
ASR errors are inevitable.
However, when a transcribed word is wrong, some subword units in the word may  still be correctly transcribed.
Therefore, enhancing the word embedding with subword units may mitigate the effect of ASR errors.

\begin{figure}[t]
  \centering
  \includegraphics[width=\linewidth]{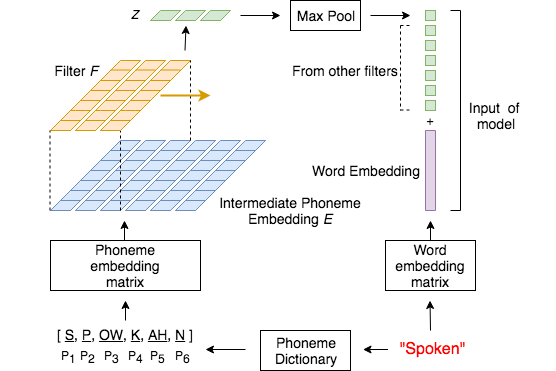}
  \caption{Illustration of enhanced word embedding. 
  For a given input word $W$ at the bottom, a sequence of phonemes $P = [p_1,...,p_l]$ are obtained by looking up in the pronunciation dictionary. 
  Each phoneme is mapped to a vector $ \mathbb{R}^{d}$ and concatenated to form intermediate matrix $E$.
  $E$ is fed into the temporal convolutional module. 
  The output $F$ is further fed into max-pooling layer, and a scalar value is generated. All the scalars from various filters will form the phoneme sequence embedding. 
  Then the phoneme sequence embedding is further concatenated with word embedding as the input  of reading comprehension model.  
  In this illustration,  $E \in \mathbb{R}^{6 \times 7}$, $F \in \mathbb{R}^{3 \times 7}$ and stride is 1.}
  \label{fig:phonemeCNN}
\end{figure}

Phoneme sequences of words are used here.
We adopt CNN to generate the  embeddings from the phoneme sequence of a word, and this network is called Phoneme-CNN. 
Our proposed approach is the reminiscent of Char-CNN~\cite{zhang2015character,kim2016character}, which apply CNN on characters to generate distributed representation of word for text classification task. 
Phoneme-CNN is illustrated in Figure~\ref{fig:phonemeCNN}.
We explain how we obtain feature for one word with one filter. 
Suppose that a word $W$ consists of a sequence of phonemes $P = [p_1,...,p_l]$, where $l$ is the number of phonemes of this word. 
Let $H \in \mathbb{R}^{C \times d}$ be the lookup table phoneme embedding matrix, where $C$ is the number of  phonemes, and $d$ is the dimension  of the phoneme embedding. 
In other words, each phoneme corresponds to a $d$-dimensional vector.
After looking up table, the embeddings of the phonemes in the word form the intermediate phoneme embedding $E \in \mathbb{R}^{l \times d}$. 
The convolution between $E$ and a filter $F \in \mathbb{R}^{k \times d}$ is performed with stride 1 to obtain one-dimension vector $Z \in \mathbb{R}^{l-k+1}$. 
After max pooling over $Z$, we obtain a scalar value. 
Since we concatenate all the output scalars from different filters, the number of filters determine the size of phoneme sequence embedding. 
The filter is essentially scanning phoneme n-gram, where the size of n-gram is the height of the filter (the number of $k$ above). 
All the parameters of filters  and phoneme embedding matrix $H$ are end-to-end learned with reading comprehension model. 
It  is also possible to incorporate other sub-word units like syllable~\cite{luong2013better,botha2014compositional,bian2014knowledge} by the same CNN architecture described above. We will experiment with syllable sequences of words.



\section{Experiments Setup and Results}
In this section, we first show the performance  of the published models from SQuAD leader board on this Spoken SQuAD dataset. 
Then we compare the performance  of models trained on text or ASR transcriptions. 
Following that with subword sequence embedding ablation study on BiDAF, we verify that phoneme sequence embedding and syllable sequence embedding improved BiDAF performance on testing sets with different WERs. 
We also provide qualitative analysis to show how typical QA model suffers from ASR errors and how phoneme sequence embedding prevent modes from these errors on spoken SQuAD dataset. 

\begin{table}[]
\centering
\caption{Experiment results for state-of-the-art models demonstrating degrading performance under spoken data. All models were trained on the full  SQuAD training set. Mnemonic Reader and FusionNet are denoted by Mreader and F-NET, respectively.  SQuAD dev set and Spoken SQuAD testing set are denoted by SQuAD-dev and SpokenS-test, respectively.}
\label{tab:stateoftheart}
\begin{tabular}{|c|c|c|c|c|}
\hline
\multicolumn{1}{|c|}{\multirow{2}{*}{\textbf{MODEL}}} &
\multicolumn{2}{|c|}{\textbf{SQuAD-dev}} & \multicolumn{2}{|c|}{\textbf{SpokenS-test}}  \\
 \cline{2-3}\cline{4-5}
\multicolumn{1}{|c|}{} & EM & F1 & EM & F1 \\
\hline
\hline
BiDAF & 58.4 & 69.9 & 37.02 & 50.9   \\
R-NET  &66.34 & 76.20 & 44.75 & 58.68 \\
Mreader & 64.00 & 73.35 &  40.36 & 52.87   \\
Dr.QA & 62.84 & 73.74 & 41.16 & 54.51\\
F-Net & 70.47 & 79.51 & 46.51 & 60.06 \\
\hline
Average & 64.41 & 74.54 & 41.96 & 55.40 \\
\hline
\end{tabular}
\end{table}

\begin{table}[]
\centering
\caption{ Performance comparison of training on  text documents (full SQuAD Train Set, denoted as \textit{Text}) and ASR transcriptions (Spoken SQuAD Train Set,  denoted as \textit{Speech}). The testing data is from Spoken SQuAD testing set (SpokenS-test) with ASR errors.} 
\label{tab:bidaf&drqa}
\begin{tabular}{|c|c|c|c|c|c|}
\hline
\multicolumn{1}{|c|}{\multirow{2}{*}{\textbf{Data}}} &
\multicolumn{1}{|c|}{\multirow{2}{*}{\textbf{Score}}} &
\multicolumn{2}{|c|}{\textbf{BiDAF}} & 
\multicolumn{2}{|c|}{\textbf{Dr.QA}}  \\
\cline{3-4}\cline{5-6}
\multicolumn{1}{|c|}{} & \multicolumn{1}{|c|}{} &  \textit{Text} &  \textit{Spoken} &  \textit{Text} &  \textit{Spoken}   \\
\hline
\hline
\multicolumn{1}{|c|}{\multirow{2}{*}{SpokenS-test}} & EM & 37.02 & 44.45 & 41.16 & 49.07   \\
\multicolumn{1}{|c|}{} & F1 & 50.9 & 57.6 & 54.51 & 61.16  \\
\hline
\end{tabular}
\end{table}

\begin{table}[]
\centering
\caption{Comparison experiments demonstrating that the proposed  approaches improved EM/F1 scores over ASR transcriptions. 
All experiments were conducted with BiDAF, which originally take word and character sequence embedding as inputs.}
\label{tab:bidaf_improve}
\begin{tabular}{|c|c|c|c|}
\hline
\multicolumn{1}{|c|}{{\textbf{MODEL}}} & \multicolumn{1}{|c|}{} &\textbf{EM} & \textbf{F1}  \\
\hline
\hline
WORD+CHAR & (a)  & 37.02 & 50.9 \\
WORD+CHAR+Dropout & (b) & 38.83 & 53.07 \\
WORD+PHONEME+Dropout& (c)    & \textbf{39.82} & \textbf{53.76} \\
WORD+SYLLABLE+Dropout & (d) & 39.71 & 53.72 \\
\hline
\end{tabular}
\end{table}

\subsection{Investigating the Impact of ASR Errors}
We trained five reading comprehension models mentioned in Section 3.1 on the SQuAD training set, and they were tested on SQuAD dev set and Spoken SQuAD testing set. 
The number of questions in the Spoken SQuAD testing set is less than the original SQuAD dev set because some of the examples are removed as mentioned in Section 2.1.
To make the comparison fair, on SQuAD dev set, we only report the results on the questions also in Spoken SQuAD testing set.
In Table~\ref{tab:stateoftheart}, across the five models, average F1 score on the text document is 75.42\%.
The average F1 score fell to 55.4\%  when there are ASR errors. 
Similar phenomenon is observed on EM.
The impact of ASR errors is significant for machine comprehension models. 

\begin{table*}[]
\centering
\caption{Performance comparison of of BiDAF with various embeddings over Spoken SQuAD testing set}
\label{tab:BIDAF}
\begin{tabular}{|c|c|c|c|c|c|c|c|c|c|c|c|c|c|c|}
\hline
\multicolumn{1}{|c|}{\multirow{2}{*}{\textbf{MODEL}}} &
\multicolumn{1}{|c|}{\multirow{2}{*}{\textbf{}}} &
\multicolumn{3}{|c|}{\textbf{No noise}} & \multicolumn{3}{|c|}{\textbf{NoiseV1}} & \multicolumn{3}{|c|}{\textbf{NoiseV2}} \\
 \cline{3-5}\cline{6-8}\cline{9-11}
\multicolumn{1}{|c|}{} & \multicolumn{1}{|c|}{} & EM & F1 & AOS & EM & F1 & AOS & EM & F1 & AOS \\
\hline
\hline
 WORD & (a) &44.34 & 57.37 & 0.3775 & 28.64 & 42.35 & 0.2915 & 19.82 &32.89 & 0.2287\\
  WORD+CHAR & (b) & 44.45 & 57.6 & 0.3772 & 29.28 &43.21 & 0.2922 & 20.07 & 33.16 & 0.2258 \\
 WORD+PHONEME & (c) &\textbf{45.58} & \textbf{58.25}& 0.3818 & 29.09 & \textbf{43.56}& 0.2899 & \textbf{20.31} & \textbf{33.42}& 0.2253\\
  WORD+SYLLABLE & (d)  &\textbf{45.61} & \textbf{58.25} & 0.3824 & \textbf{29.37} & \textbf{43.46} & 0.2974 & \textbf{20.23} & \textbf{33.53} & 0.2316\\
\hline
 WORD+CHAR+PHONEME+SYLLABLE & (e) & \textbf{45.78} & \textbf{58.71}& 0.3846 & \textbf{29.73} & \textbf{44.2}& 0.2967 & \textbf{20.66} & \textbf{33.86}& 0.2295 \\
\hline
\end{tabular}
\end{table*}   

\subsection{Training on ASR Transcriptions}
We further trained BiDAF and Dr.QA on either  SQuAD training set (text documents) or Spoken SQuAD training set (ASR transcriptions), and tested on the testing set of Spoken SQuAD. 
The results are in Table~\ref{tab:bidaf&drqa}. 
The results for training on text docuemtns are labelled with  \textit{Text}, while training on ASR transcriptions are labelled as \textit{Speech}.
The scale of SQuAD training data is almost two times of Spoken SQuAD training data because some documents were removed\footnote{In these documents, the correct answers do not exist in the ASR transcriptions due to ASR errors, so they cannot be used to train BiDAF and Dr.QA.}. 
However, when testing documents have ASR errors,   models trained on ASR transcriptions are remarkably better than on text documents. 


\subsection{Mitigating ASR errors by Subword Units}
We utilized CMU LOGIOS Lexicon Tool to convert each word into sequence of phonemes and then fed the phonemes into Phoneme-CNN network to obtain phoneme sequence embedding.  The network details are listed as follow : phoneme embedding size 6, filter size 3x6 and numbers of filters 80.  Different from~\cite{li2016phoneme} using one-hot vector, we choose distributed representation vectors to represent phonemes.
For syllables, we utilized a python module Pyphen that hyphenate text uses existing Hunspell hyphenation dictionaries to convert word to sequence of syllables. It is reported that in English, hyphenation output of word is often the same as sequence of syllables in that word. 
We fed the syllables into Phoneme-CNN network to obtain the syllable sequence embedding. The network details are listed as follow : syllable embedding size 20, filter size 2x20 and numbers of filters 100. Both phoneme sequence embedding and syllable sequence embedding will be further concatenated with other representations of word to be the inputs of reading comprehension model. 

The experimental results with various mitigation approaches which were trained on SQuAD training set are in Table~\ref{tab:bidaf_improve}. We chose  BiDAF as the reading comprehension model. 
Because ASR errors can be considered as noise, we can see that adding dropout offered improvements (rows (b) v.s. (a)), while using phoneme or syllable sequence embedding is even better (rows (c), (d) v.s. (b)). 

We compare the performance of the proposed subword unit sequence embedding with two strong baselines, word embedding and combination of word and character embedding. 
The results on Spoken SQuAD testing set without noise and different levels of noises are listed in Table~\ref{tab:BIDAF}. 
We see from Table~\ref{tab:BIDAF}, phoneme and syllable sequence embedding mostly performed better (row (c)(d) vs. (a)(b)). 
In addition, model which concatenates character, phoneme and syllable sequence embeddings together (row (e)) achieves the best performance across EM/F1/AOS over all kinds of WER ASR data.  


\subsection{Qualitative Analysis}
Table~\ref{tab:EXAMPLE} is a selected example from Spoken SQuAD testing set.
According to the generated predictions, we observe that BiDAF with only word embedding is not robust to ASR errors.  
We can check on ASR transcriptions, the word "area" is misrecognized to "harry". 
The BiDAF model with only word embedding is confused by the word "harry". The corresponding question starts with "Who", so the model is trying to search for a person name neighboring the key word "Los Angeles county". However, model with our proposed phoneme sequence embedding correctly selects the answer span despite of the ASR transcription errors.

\begin{table}[]
\centering
\caption{An example of generated predictions from BiDAF with proposed phoneme sequence embedding and BiDAF with only word embedding. The capitalization was added on ASR transcriptions for easy reading.} 
\label{tab:EXAMPLE}
\begin{tabular}{|c p{4cm}|}
\hline
\multirow{7}{*}{\textbf{Text Document}}  &  To the east, the United States Census Bureau considers the San Bernardino and Riverside County areas, Riverside-San Bernardino area as a separate metropolitan area from Los Angeles County.\\
\hline
\multirow{7}{*}{\textbf{ASR Transcription}}  & To the east, the United States Census Bureau considers the San Bernardino and Riverside County areas, Riverside San Bernardino \textbf{harry} as a separate separate metropolitan area from Los Angeles county.\\
\hline
\multirow{3}{*}{\textbf{Question}} & Who considers Los Angeles County to be a separate metropolitan area? \\
\hline
\textbf{Ground truth} & United States Census Bureau  \\
\hline
\textbf{WORD} in Table~\ref{tab:BIDAF} & riverside san bernardino harry \\
\hline
\textbf{WORD+PHONEME}& \multirow{2}{*}{United States Census Bureau}   \\
in Table~\ref{tab:BIDAF} & \\
\hline
\end{tabular}
\end{table}

\section{Conclusions}
In this paper we propose a new extraction based spoken question answering task named Spoken SQuAD. 
We demonstrate that ASR errors significantly degraded the performance of reading comprehension models. Then we propose to use different kinds of subword units to mitigate the impact of ASR errors.

\bibliographystyle{IEEEtran}

\bibliography{mybib}

\end{document}